\pdfoutput=1

\def\year{2019}\relax

\documentclass[letterpaper]{article}    % DO NOT CHANGE THIS
\usepackage{aaai19}                     % Required
\usepackage{times}                      % Required
\usepackage{helvet}                     % Required
\usepackage{courier}                    % Required
\usepackage{url}                        % Required
\usepackage{graphicx}                   % Required
\usepackage{verbatim}
\usepackage{amsmath}
\usepackage{amssymb}
\usepackage[ruled]{algorithm2e}
\usepackage{cuted}
\usepackage{capt-of}
\usepackage{enumitem}
\usepackage{acronym}
\usepackage{xspace}
\usepackage{subfig}

\acrodef{mrf}[MRF]{Markov random field}
\acrodef{maml}[MAML]{model-agnostic meta-learning}

\title{MetaStyle: Three-Way Trade-Off Among\\Speed, Flexibility, and Quality in Neural Style Transfer}
\author{Chi Zhang \and Yixin Zhu \and Song-Chun Zhu \\
\texttt{\{chizhang,yzhu,sczhu\}@cara.ai}\\
International Center for AI and Robot Autonomy (CARA)
}

\DeclareMathOperator*{\argmin}{arg\,min}
\newcommand{\norm}[1]{\left\lVert#1\right\rVert}

\makeatletter
\DeclareRobustCommand\onedot{\futurelet\@let@token\@onedot}
\def\@onedot{\ifx\@let@token..\else.\null\fi\xspace}
\def\etal{\emph{et al}\onedot}
\def\eg{\emph{e.g}\onedot} 
\def\ie{\emph{i.e}\onedot} 
 
\makeatother

\begin{document}

\maketitle

% \begin{strip}
%     \includegraphics[width=\linewidth]{figures/prologue}
%     \captionof{figure}{Style transfer results using MetaStyle, balancing the three-way trade-off among speed, flexibility, and quality. Left: the content image and the style-free representation learned by MetaStyle. Right: the stylized images from 14 different styles.\label{fig:prologue}}
% \end{strip}

\begin{abstract}
An unprecedented booming has been witnessed in the research area of artistic style transfer ever since Gatys \etal introduced the neural method. One of the remaining challenges is to balance a trade-off among three critical aspects---speed, flexibility, and quality: (i) the vanilla optimization-based algorithm produces impressive results for arbitrary styles, but is unsatisfyingly slow due to its iterative nature, (ii) the fast approximation methods based on feed-forward neural networks generate satisfactory artistic effects but bound to only a limited number of styles, and (iii) feature-matching methods like AdaIN achieve arbitrary style transfer in a real-time manner but at a cost of the compromised quality. We find it considerably difficult to balance the trade-off well merely using a single feed-forward step and ask, instead, whether there exists an algorithm that could adapt quickly to any style, while the adapted model maintains high efficiency and good image quality. Motivated by this idea, we propose a novel method, coined \emph{MetaStyle}, which formulates the neural style transfer as a bilevel optimization problem and combines learning with only a few post-processing update steps to adapt to a fast approximation model with satisfying artistic effects, comparable to the optimization-based methods for an arbitrary style. The qualitative and quantitative analysis in the experiments demonstrates that the proposed approach achieves high-quality arbitrary artistic style transfer effectively, with a good trade-off among speed, flexibility, and quality.
\end{abstract}

\begin{figure*}[t!]
    \centering
    \includegraphics[width=\linewidth]{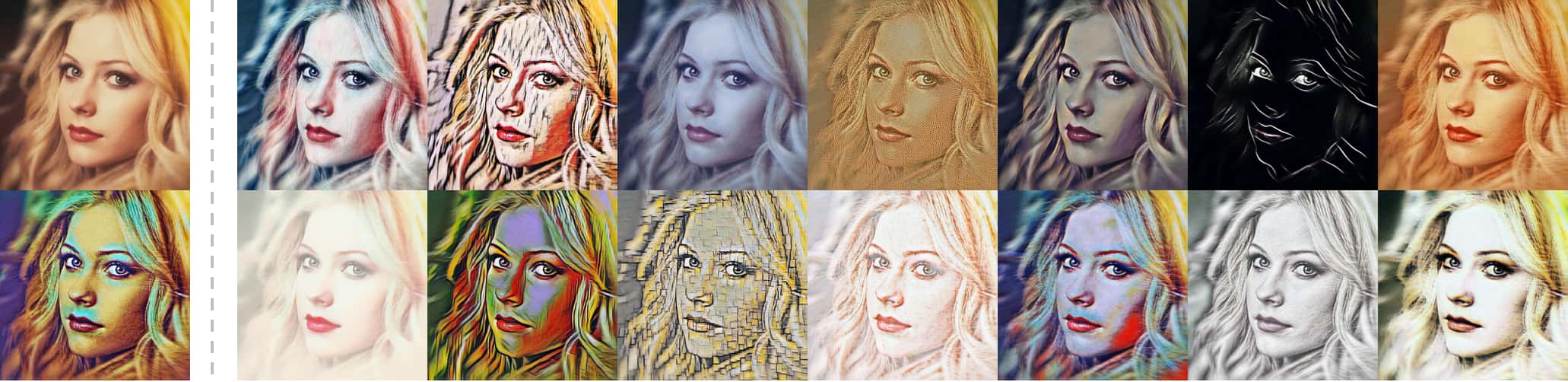}
    \caption{Style transfer results using MetaStyle, balancing the three-way trade-off among speed, flexibility, and quality. Left: the content image and the style-free representation learned by MetaStyle. Right: the stylized images from 14 different styles}
    \label{fig:prologue}
\end{figure*}

\section{Introduction}

To reduce the strenuous early-day efforts in producing pastiche, the computer vision and machine learning community have joined forces to devise automated algorithms to render a content image in the same style from a source artistic work. The style transfer problem covers a wide range of work, and at the beginning was phrased as a texture synthesis \cite{diaconis1981statistics,zhu1998filters} problem. Some notable work includes: (i) non-parametric sampling methods \cite{efros1999texture} and acceleration methods by a tree-structured vector quantization \cite{wei2000fast}, (ii) patch-based sampling methods \cite{efros2001image,liang2001real} for better quality and efficiency, (iii) energy minimization methods using EM-like algorithms \cite{kwatra2005texture}, and (iv) image analogies \cite{hertzmann2001image} to produce the ``filtered'' results and their extensions to portrait paintings \cite{zhao2011portrait}. % \citeauthor{efros1999texture} \shortcite{efros1999texture} first proposed to solve this problem by growing texture pixels one by one outward using a non-parametric sampling, and \citeauthor{wei2000fast} \shortcite{wei2000fast} accelerated this process by a tree-structured vector quantization. Patch-based sampling methods \cite{efros2001image,liang2001real} were later proposed to improve the synthesis quality and efficiency. \citeauthor{kwatra2005texture} \shortcite{kwatra2005texture}, however, viewed the problem from an energy minimization perspective and jointly optimized the objective using an EM-like algorithm. The concept of image analogies \cite{hertzmann2001image} was also introduced to produce the ``filtered'' results and later extended by \citeauthor{zhao2011portrait} \shortcite{zhao2011portrait} to tailor to portrait paintings.

With the recent boost of deep neural networks and large datasets in computer vision, \citeauthor{gatys2016image} \shortcite{gatys2016image} first discovered that combining multi-level VGG features \cite{simonyan2014very} trained on the ImageNet \cite{deng2009imagenet} successfully captured the characteristics of the style while balancing the statistics of the content, producing impressive results for the task of artistic style transfer. This serendipitous finding has brought to life a surge of interests in the research area of style transfer. Iterative optimization methods \cite{gatys2015texture,gatys2016image,li2016combining} generate artistic images that well interpolate between arbitrary style space and content space; but due to its iterative nature, these methods are generally slow, requiring hundreds of update steps and becoming impractical for deployment in products. Feed-forward neural networks trained with \emph{perceptual loss} \cite{johnson2016perceptual,dumoulin2017learned,zhang2017multi} overcome the speed problem and usually result in satisfactory artistic effects; however, good quality is limited to a single or a small number of style images, sacrificing the flexibility in the original method. Feature-matching methods \cite{huang2017arbitrary,sheng2018avatar} achieve arbitrary style transfer in real-time, but these models come at the cost of compromised style transfer quality, compared to the methods mentioned above. 

To address these problems, we argue that it is nontrivial to use either sheer iterative optimization methods or single-step feed-forward approximations to achieve the three-way trade-off among speed, flexibility, and quality. In this work, we seek to find, instead, an algorithm that would fast adapt to any style by a small or even negligible number of post-processing update steps, so that the adapted model keeps high efficiency and satisfactory generation quality. 

Specifically, we propose a novel style transfer algorithm, coined \emph{MetaStyle}, which formulates the fast adaptation requirement as the bilevel optimization, solvable by the recent meta-learning methods \cite{finn2017model,nichol2018on}. This unique problem formulation encourages the model to learn a style-free representation for content images, and to produce a new feed-forward model, after only a small number of update steps, to generate high-quality style transfer images for a single style efficiently. From another perspective, this formulation could also be thought of as finding a style-neutral input for the vanilla optimization-based methods \cite{gatys2016image}, but transferring styles much more effectively.

Our model is instantiated using a neural network. The network structure is inspired by the finding \cite{dumoulin2017learned} that scaling and shifting parameters in instance normalization layers \cite{ulyanov2017improved} are specialized for specific styles. In contrast, unlike prior work, our method implicitly forces the parameters to find no-style features in order to rapidly adapt and remain parsimonious in terms of the model size. The trained MetaStyle model has roughly the same number of parameters as described in \citeauthor{johnson2016perceptual} \shortcite{johnson2016perceptual}, and requires merely 0.1 million training steps.

Comprehensive experiments with both qualitative and quantitative analysis, compared with prior neural style transfer methods, demonstrate that the proposed method achieves a good trade-off among speed, flexibility, and quality. Figure~\ref{fig:prologue} shows sample results using the proposed style transfer.

The contributions of the paper are three-fold:
\begin{itemize}[leftmargin=*,noitemsep,nolistsep]
    \item We propose a new style transfer method called MetaStyle to achieve the three-way trade-off in speed, flexibility, and quality. To the best of our knowledge, this is the first paper that formulates the style transfer as the bilevel optimization so that the model could be easily adapted to a new style with only a small number of updates, producing high-quality results while remaining parsimonious.
    \item The proposed method provides a style-free representation, from which a fast feed-forward high-quality style transfer model could be adapted after only a small number of iterations, making the cost of training a high-quality model for a new style nearly negligible.
    \item The proposed method results in a style-neutral representation that comes with better convergence for vanilla optimization-based style transfer methods.
\end{itemize}

\section{Related Work}

\subsection{Neural Style Transfer}

By leveraging the pre-trained VGG model \cite{simonyan2014very}, \citeauthor{gatys2016image} \shortcite{gatys2016image} first proposed to explicitly separate content and style: the model has a feature-matching loss involving the second-order Gram matrices (later called \emph{perceptual loss}) and iteratively updates the input images (usually hundreds of iterations) to produce high-quality style transfer results. To overcome the speed limit, \citeauthor{johnson2016perceptual} \shortcite{johnson2016perceptual} recruited an image transformation network to generate stylized results sufficiently close to the optimum solution directly. Concurrent work by \citeauthor{ulyanov2016texture} \shortcite{ulyanov2016texture} instantiated a similar idea using multi-resolution generator network and further improved the diversity of the generated images \cite{ulyanov2017improved} by applying the Julesz ensembles \cite{zhu1998filters,zhu2000exploring}. Note that each trained model using any of these methods is specialized to a single style.

Significant efforts have been made to improve the neural style transfer. \citeauthor{li2016combining} \shortcite{li2016combining} modeled the process using an \ac{mrf} and introduced the \ac{mrf} loss for the task. \citeauthor{li2017demystifying} \shortcite{li2017demystifying} discovered that the training loss could be cast in the maximum mean discrepancy framework and derived several other loss functions to optimize the content image. \citeauthor{chen2017stylebank} \shortcite{chen2017stylebank} jointly learned a style bank for each style during model training. \citeauthor{dumoulin2017learned} \shortcite{dumoulin2017learned} modified the instance normalization layer \cite{ulyanov2017improved} to condition on each style. \citeauthor{zhang2017multi} \shortcite{zhang2017multi} proposed to use a CoMatch layer to match the second-order statistics to ease the learning process. Although these approaches produce transfer results of good quality in real-time for a constrained set of styles, they still lack the generalization ability to transfer to arbitrary styles. Additionally, these approaches sometimes introduce additional parameters proportional to the number of the styles they learn.

Recent work concentrated on more generalizable approaches. A patch-based style swap layer was first introduced \cite{chen2016fast} to replace the content feature patch with the closest-matching style feature patch, and a compromised inverse network was employed for fast approximation. The adaptive instance normalization layer \cite{huang2017arbitrary} was introduced to scale and shift the normalized content features by style feature statistics and act as the bottleneck in an encoder-decoder architecture, while similarly \citeauthor{li2017universal} \shortcite{li2017universal} applied recursive whitening and coloring transformation in multi-level pre-trained auto-encoder architecture. More recent works include a ZCA-like style decorator and an hourglass network that were integrated in a multi-scale manner \cite{sheng2018avatar} and a meta network that was trained to generate parameters of an image transformation network \cite{shen2018neural} directly. These methods, though efficient and flexible, often suffer from compromised image generation quality, especially for the unobserved styles. In contrast, the proposed model could adapt to any style quickly without sacrificing the speed or the image quality on par with fast approximation methods, \eg, \citeauthor{johnson2016perceptual} \shortcite{johnson2016perceptual}.

Additionally, our model is also parsimonious, requiring roughly the same number of model parameters as \citeauthor{johnson2016perceptual} \shortcite{johnson2016perceptual}, using merely 0.1 million iterations. In comparisons, \eg, \citeauthor{ghiasi2017exploring} \shortcite{ghiasi2017exploring} extended the conditional instance normalization framework \cite{dumoulin2017learned}, but required a pre-trained Inception-v3 \cite{szegedy2016rethinking} to predict the parameters for a single style. This model requires 4 million update steps, making training burdensome.

\subsection{Meta-Learning}

Meta-learning has been successfully applied in few-shot learning with early work dated back to the 1990's. Here we only review one branch focusing on \emph{initialization strategy} \cite{franceschi18bilevel} that influences our work. \citeauthor{ravi2016optimization} \shortcite{ravi2016optimization} first employed an LSTM network as a meta-learner to learn an optimization procedure. \citeauthor{finn2017model} \shortcite{finn2017model} proposed \ac{maml} so that a model previously learned on a variety of tasks could be quickly adapted to a new one. This method, however, required second-order gradient computation in order to derive gradient for the meta-objective correctly, and therefore consumed significant computational power, though a first-order method was also tested with compromised performance.

Following their work, \citeauthor{nichol2018on} \shortcite{nichol2018on} generalized \ac{maml} to a family of algorithms and extended it to Reptile. Reptile coupled sequential first-order gradients with advanced optimizers, such as Adam \cite{kingma2014adam}, resulting in an easier implementation, shorter training time and comparable performance. A recent work \cite{shen2018neural} modeled the process of neural style transfer using an additional large group of fully-connected layers such that the parameters of an image transformation network could be predicted. In contrast, the proposed method remains parsimonious with a single set of parameters to train and adapt.

As we will show in the Section~\ref{sec:formulate}, the meta network is, \emph{de facto}, a special case in the proposed bilevel optimization framework. To the best of our knowledge, our paper is the first to explicitly cast neural style transfer as the bilevel optimization problem in the initialization strategy branch.

\section{Background}

Before detailing the proposed model, we first introduce two essential building blocks, \ie, the perceptual loss and the general bilevel optimization problem, which lay the foundation of the proposed approach.

\subsection{Style Transfer and Perceptual Loss}

Given an image pair $(I_c, I_s)$, the style transfer task aims to find an ``optimal'' solution $I_x$ that preserves the content of $I_c$ in the style of $I_s$. \citeauthor{gatys2016image} \shortcite{gatys2016image} proposed to measure the optimality with a newly defined loss using the trained VGG features, later modified and named as the perceptual loss \cite{johnson2016perceptual}. The perceptual loss could be decomposed into two parts: the content loss and the style loss.

Denoting the VGG features at layer $i$ as $\phi_i(\cdot)$, the content loss $\ell_{\text{content}}(I_c, I_x)$ is defined using the $L_2$ norm
\begin{equation}
    \ell_{\text{content}}(I_c, I_x) = \frac{1}{N_i} \norm{\phi_i(I_c) - \phi_i(I_x)}_2^2,
\end{equation}
where $N_i$ denotes the number of features at layer $i$.

The style loss $\ell_{\text{style}}(I_s, I_x)$ is the sum of Frobenius norms between the Gram matrices of the VGG features at different layers
\begin{equation}
    \ell_{\text{style}}(I_s, I_x) = \sum_{i \in S} \norm{G(\phi_i(I_s)) - G(\phi_i(I_x))}_F^2,
\end{equation}
where $S$ denotes a predefined set of layers and $G$ the Gramian transformation.

The transformation could be efficiently computed by 
\begin{equation}
    G(x) = \frac{\psi(x) \psi(x)^T}{C H W}
\end{equation}
for a 3D tensor $x$ of shape $C \times H \times W$, where $\psi(\cdot)$ reshapes $x$ into $C \times H W$.

The perceptual loss $\ell(I_c, I_s, I_x)$ aggregates the two components by the weighted sum
\begin{equation}
    \ell(I_c, I_s, I_x) = \alpha \ell_{\text{content}}(I_c, I_x) + \beta \ell_{\text{style}}(I_s, I_x).
\end{equation}

\subsection{Bilevel Optimization}
\label{sec:bilevel}

We formulate the style transfer problem as the bilevel optimization in the form simplified by \citeauthor{franceschi18bilevel} \shortcite{franceschi18bilevel}
\begin{equation}
    \begin{aligned}
        & \underset{\theta}{\text{minimize}}    & & E(w_\theta, \theta) \\
        & \text{subject to}                     & & w_\theta = \argmin_w L_\theta(w),
    \end{aligned}
\end{equation}
where $E$ is the \emph{outer objective} and $L_\theta$ the \emph{inner objective}. Under differentiable $L_\theta$, the constraint could be replaced with $\nabla L_\theta = 0$. However, in general, no closed-form solution of $w_\theta$ exists and a practical approach to approximate the optimal solution is to replace the inner problem with the gradient dynamics, \ie,
\begin{equation}
    \begin{aligned}
        & \underset{\theta}{\text{minimize}}    & & E(w_T, \theta)      \\
        & \text{subject to}                     & & w_0 = \Psi(\theta)  \\
        &                                       & & w_{t} = w_{t - 1} - \delta \nabla L_\theta(w_{t - 1})
    \end{aligned}
    \label{prob:relaxed}
\end{equation}
where $\Psi$ initializes $w_0$, $\delta$ is the step size and $T$ the maximum number of steps. \citeauthor{franceschi18bilevel} \shortcite{franceschi18bilevel} proved the convergence of Equation~\ref{prob:relaxed} under certain conditions. Though they did not model their problems using bilevel optimization but rather an intuitive motivation, \citeauthor{finn2017model} \shortcite{finn2017model} and \citeauthor{nichol2018on} \shortcite{nichol2018on} both use the identity mapping for $\Psi$, with the former computing the full gradient for $\theta$ to optimize the outer objective, and the latter one only the first-order approximate gradient.

\section{MetaStyle}

In this section, we first detail the intuition behind and the formulation of the proposed framework, explain the design choices and discuss relations to the previous approaches. Then the network architecture is presented with the training protocol and the detailed algorithm.

\subsection{Problem Formulation}
\label{sec:formulate}

\begin{figure}[t!]
    \centering
    \includegraphics[width=\linewidth]{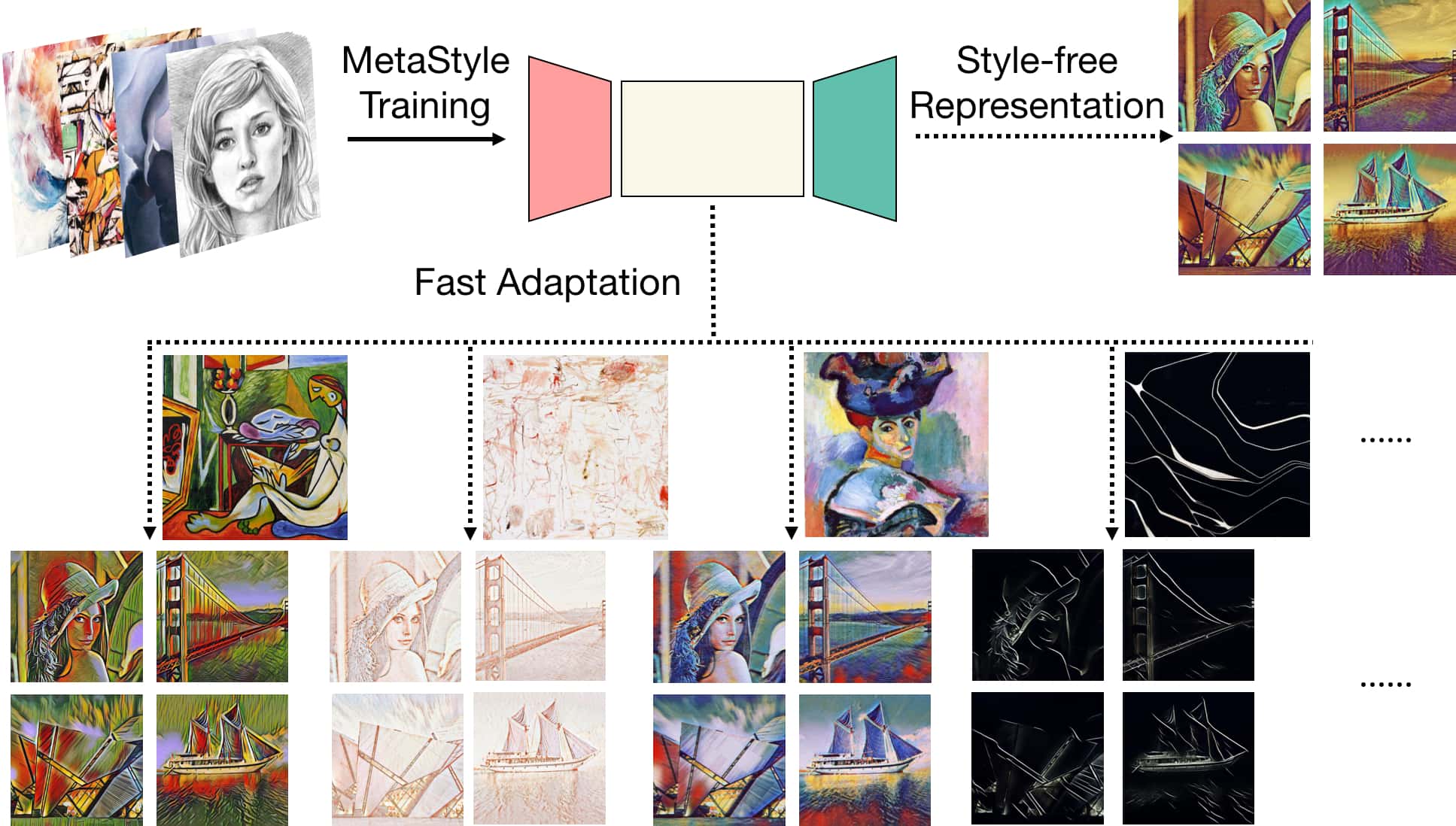}
    \caption{The proposed MetaStyle framework, in which the model is optimized using the bilevel optimization over large-scale content and style dataset. The framework first learns a style-neutral representation. A limited number of post-processing update steps is then applied to adapt the model quickly to a new style. After adaptation, the new model serves as an image transformation network with good transfer quality and high efficiency.}
    \label{fig:procedure}
\end{figure}

MetaStyle is tasked with finding a three-way trade-off among speed, flexibility, and quality in neural style transfer. To achieve such a balance, however, we argue that it is nontrivial to either merely use iterative optimization methods or simply adopt single-step feed-forward approximations. To address this challenge, we consider a new approach where we first learn a style-neutral representation and allow a limited number of update steps to this neutral representation in the post-processing stage to adapt to a new style. It is expected that the model should generate a stylized image efficiently after adaptation, be general enough to accommodate any new style, and produce high-quality results.

To this end, we employ an image transformation network with content image input \cite{johnson2016perceptual} and cast the entire neural style transfer problem in a bilevel optimization framework \cite{franceschi18bilevel}. As discussed in Equation \ref{prob:relaxed}, we choose to model $\theta$ as the network initialization and $w_T$ the adapted parameters, now denoted as $w_{s, T}$, to emphasize the style to adapt to. $T$ is restricted to be small, usually in the range between 1 and 5. Both the inner and outer objective is designed to be the perceptual loss averaged across datasets. However, as described in meta-learning \cite{finn2017model,nichol2018on}, the inner objective uses a model initialized with $\theta$ and only optimizes contents in the training set, whereas the outer objective tries to generalize to contents in the validation set. $\Psi$ is the identity mapping. Formally, the problem could be stated as

{\fontsize{8.5pt}{9pt} \selectfont
\begin{equation}
    \begin{aligned}
        &\underset{\theta}{\text{minimize}}    & & \mathbb{E}_{c,s}[\ell(I_c, I_s, M(I_c; w_{s, T}))] \\
        &\text{subject to}                     & & w_{s, 0} = \theta \\
        & & & w_{s, t} = w_{s, t - 1} - \delta \nabla \mathbb{E}_c[\ell(I_c, I_s, M(I_c; w_{s, t - 1}))],
    \end{aligned}
    \label{prob:formulate}
\end{equation}
}where $M(\cdot; \cdot)$ denotes our model and $\delta$ the learning rate of the inner objective. The expectation of the outer objective $\mathbb{E}_{c,s}$ is taken with respect to both the styles and the content images in the validation set, whereas the expectation of the inner objective $\mathbb{E}_c$ is taken with respect to the content images in the training set only. This design allows the adapted model to specialize for a single style but still maintain the initialization generalized enough. Note that for the outer objective, $w_{s, T}$ implicitly depends on $\theta$. In essence, the framework learns an initialization $M(\cdot; \theta)$ that could adapt to $M(\cdot; w_{s, T})$ efficiently and preserve high image quality for an arbitrary style. Figure~\ref{fig:procedure} shows the proposed framework.

The explicit training-validation separation in the framework forces the style transfer model to generalize to unobserved content images without over-fitting to the training set. Coupled with this separation, MetaStyle constrains the number of steps in the gradient dynamics computation to encourage quick adaptation for an arbitrary style and, at the same time, picks an image transformation network due to its efficiency and high transfer quality. These characters serve to the trade-offs among speed, flexibility, and quality.

We now discuss MetaStyle's relations to other methods.

\textbf{Relation to Johnson \etal \shortcite{johnson2016perceptual}:}
Johnson \etal's method finds an image transformation model tailored to a single given style, minimizing the model parameters by
\begin{align}
    \underset{w}{\text{minimize}} \quad \mathbb{E}_c[\ell(I_c, I_s, M(I_c; w))],
    \label{prob:jc}
\end{align}
where the expectation is taken with respect to only the contents. In contrast, in Equation~\ref{prob:formulate}, we seek a specific model \emph{initialization} $\theta$, which is not the final parameters used for the style transfer, but could adapt to any other style using merely a small number of post-processing updates. Assuming there exists an implicit, unobserved neutral style, MetaStyle could be regarded as learning a style-free image transformation.

\textbf{Relation to Gatys \etal \shortcite{gatys2016image}:}
Starting with the content image, Gatys \etal finds the minimizer of the perceptual loss using iterative updates. From this iterative update perspective, MetaStyle could be regarded as learning to find a good starting point for the optimization algorithm. This learned transformation generates a style-neutral image while dramatically reducing the number of update steps.

\begin{figure}[t!]
    \centering
    \includegraphics[width=\linewidth]{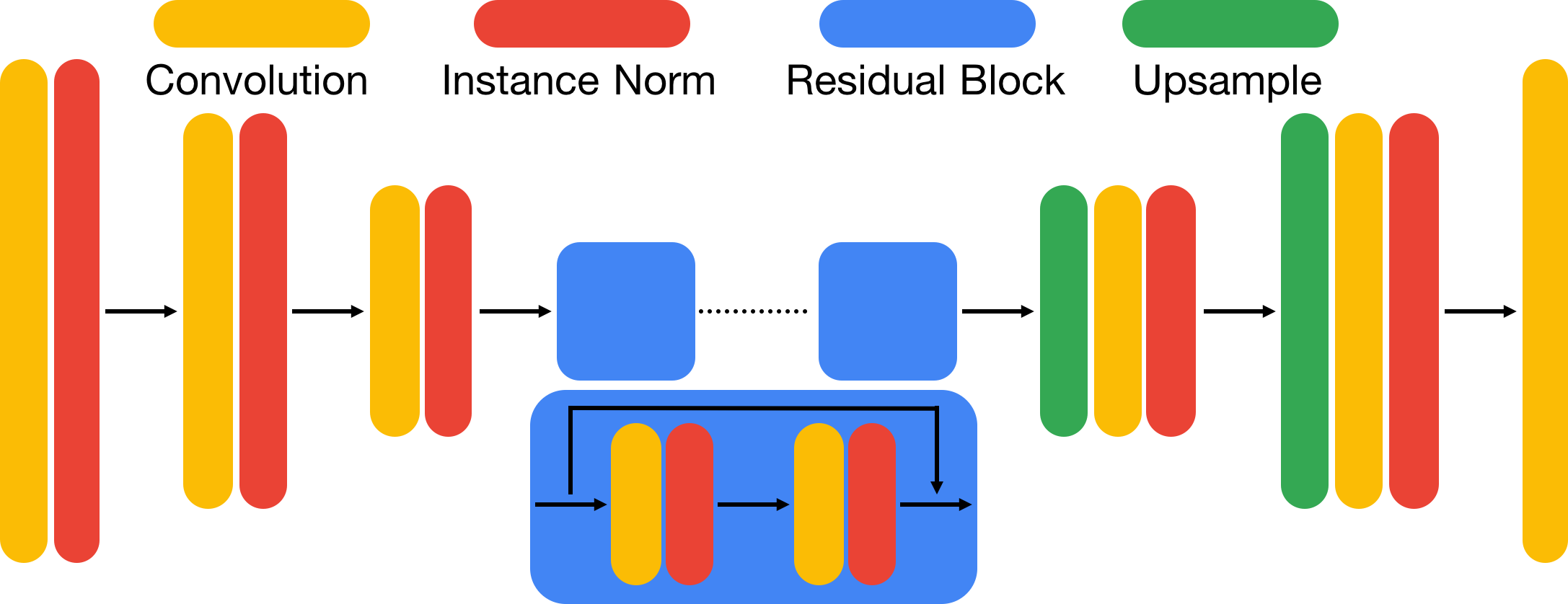}
    \caption{Network architecture. Residual Blocks are stacked multiple times to extract deeper image features.}
    \label{fig:net}
\end{figure}

\textbf{Relation to Shen \etal \shortcite{shen2018neural}:}
Shen \etal's method is a special case of the proposed bilevel optimization framework, where $T = 0$ and $\Psi$ is a highly nonlinear transformation, parameterized by $\theta$ that uses a style image to predict parameters of another image transformation network.

\begin{algorithm}[t!]
    \DontPrintSemicolon
    \linespread{1.2}\selectfont
    \SetKwInOut{Input}{Input}
    \SetKwInOut{Output}{Output}
    \Input{content training dataset $\mathcal{D}_{tr}$, content validation dataset $\mathcal{D}_{val}$, style dataset $\mathcal{D}_{style}$, inner learning rate $\delta$, outer learning rate $\eta$, number of inner updates $T$}
    \Output{trained parameters $\theta$}
    \BlankLine
    randomly initialize $\theta$\;
    \While{not done}{
        initialize outer loss $E \leftarrow 0$\;
        sample a batch of styles from $\mathcal{D}_{style}$\;
        \For{each style $I_s$}{
            $w_{s} \leftarrow \theta$\;
            \For{$i \leftarrow 1$ \KwTo $T$}{
                sample a batch $\mathcal{B}_{tr}$ from $\mathcal{D}_{tr}$\;
                compute inner loss $L_\theta$ using $I_s$ and $\mathcal{B}_{tr}$\;
                $w_{s} \leftarrow w_{s} - \delta \nabla L_\theta$\;
            }
            sample a batch $\mathcal{B}_{val}$ from $\mathcal{D}_{val}$\;
            increment $E$ by loss from $I_s$ and $\mathcal{B}_{val}$\;
        }
        $\theta \leftarrow \theta - \eta \nabla E$\;
    }
    \caption{MetaStyle}
    \label{alg:metastyle}
\end{algorithm}

\subsection{Network Architecture, Training \& Algorithm}

Our network architecture largely follows that of an image transformation network described in \citeauthor{dumoulin2017learned} \shortcite{dumoulin2017learned}. However, unlike the original architecture, the output of the last convolution layer is unnormalized and activated using the Sigmoid function to squash it into $[0, 1]$. Upsampled convolution, which first upsamples the input and then performs convolution, and reflection padding are used to avoid checkerboard effects \cite{zhang2017multi}. Inspired by the finding \cite{dumoulin2017learned} that scaling and shifting parameters in the instance normalization layers specialize for specific styles, we append an instance normalization layer after each convolution layer, except the last. See Figure~\ref{fig:net} for a graphical illustration. This design forces the parameters in instance normalization layers to learn from an implicit, unobserved neutral style while keeping the model size parsimonious.

For training, we use small-batch learning to approximate both the inner and outer objective. The inner objective is approximated by several batches sampled from the training dataset and computed on a single style, whereas the outer objective is approximated by a style batch, in which each style incurs a perceptual loss computed over a content batch sampled from the validation dataset. The problem is solvable by \ac{maml} \cite{finn2017model} and summarized in Algorithm~\ref{alg:metastyle}. After training, $\theta$ could be used as the initialization to minimize Equation~\ref{prob:jc} to adapt the model to a single style or to provide the starting point $M(I_c; \theta)$ for optimization-based methods.

\section{Experiments}

\subsection{Implementation Details}

We train our model using MS-COCO \cite{lin2014microsoft} as our content dataset and WikiArt test set \cite{nichol2016painter} as our style dataset. The content dataset has roughly 80,000 images and the WikiArt test set 20,000 images. We use Adam \cite{kingma2014adam} with a learning rate 0.001 to optimize the outer objective and vanilla SGD with a learning rate 0.0001 for the inner objective. All batches are of size 4. We fix $\alpha = 1$, $\beta = 1 \times 10^5$ across all the experiments. Content loss is computed on \verb|relu2_2| of a pre-trained VGG16 model and style loss over \verb|relu1_2|, \verb|relu2_2|, \verb|relu3_3| and \verb|relu4_3|. To encourage fast adaptation, we constrain $T = 1$. The entire model is trained on a Nvidia Titan Xp with only 0.1 million iterations.

\begin{table}[b!]
    \centering
    \begin{tabular}{l | c | c | c | c}
    \hline
    Method        & Param            & 256 (s)         & 512 (s)         & \# Styles \\
    \hline
    Gatys \etal   & N/A              & 7.7428          & 27.0517         & $\infty$ \\
    Johnson \etal & \textbf{1.68M}   & \textbf{0.0044} & \textbf{0.0146} & 1        \\
    Li \etal      & 34.23M           & 0.6887          & 1.2335          & $\infty$ \\
    Huang \etal   & 7.01M            & 0.0165          & 0.0320          & $\infty$ \\
    Shen \etal    & 219.32M          & \textbf{0.0045} & \textbf{0.0147} & $\infty$ \\
    Sheng \etal   & 147.22M          & 0.5089          & 0.6088          & $\infty$ \\
    Chen \etal    & \textbf{1.48M}   & 0.2679          & 1.0890          & $\infty$ \\
    \hline
    \textbf{Ours} & \textbf{1.68M}   & \textbf{0.0047} & \textbf{0.0145} & $\infty^\star$ \\
    \hline
    \end{tabular}
    \caption{Speed and flexibility benchmarking results. Param lists the number of parameters in each model. 256/512 denotes inputs of $256 \times 256$/$512 \times 512$. \# Styles represents the number of styles a model could potentially handle. $^\star$Note that MetaStyle adapts to a specific style after very few update steps and the speed is measured for models adapted.}
    \label{tbl:benchmark}
\end{table}

\subsection{Comparison with Existing Methods}

\begin{figure*}[t!]
    \centering
    \includegraphics[width=\linewidth]{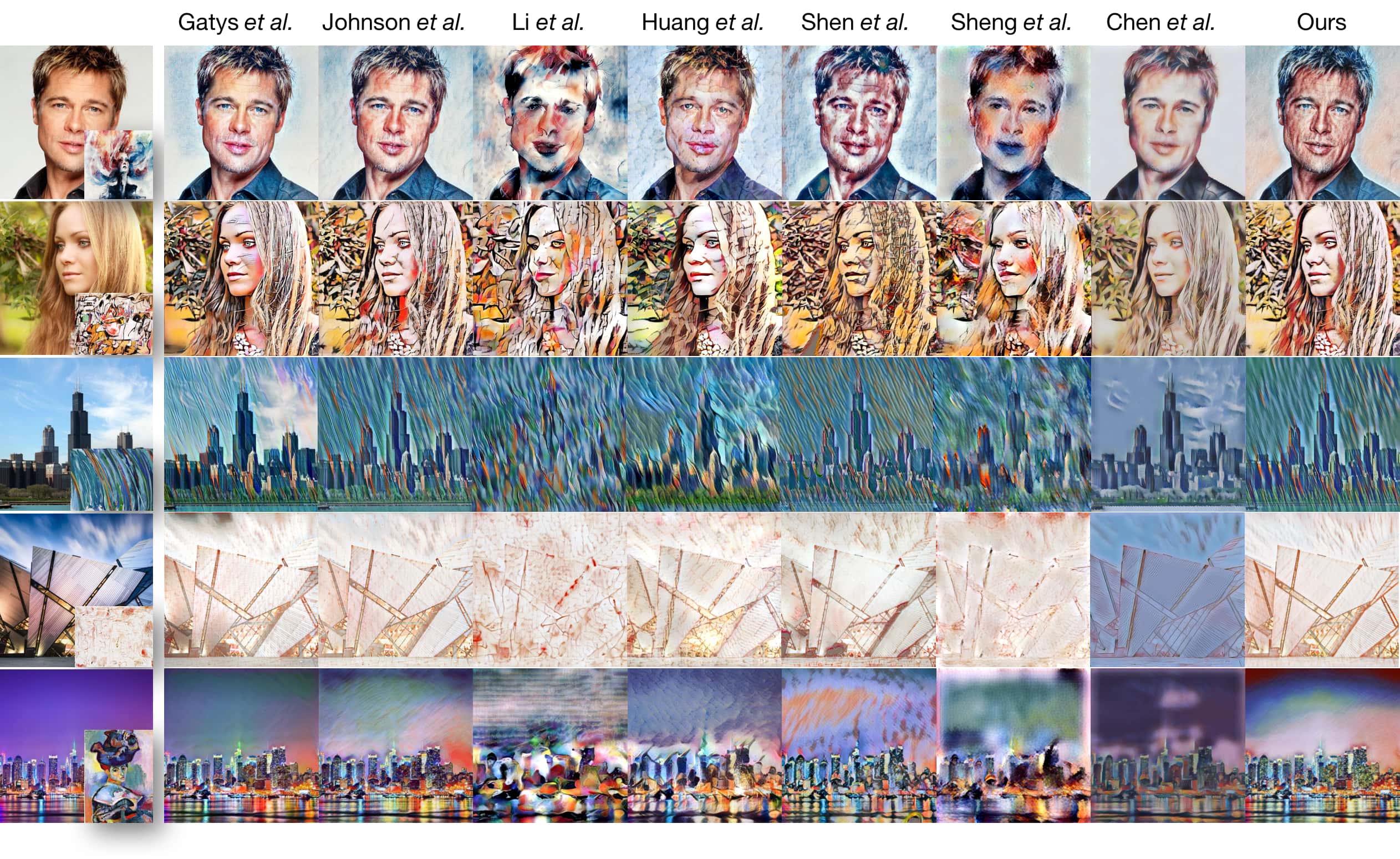}
    \caption{Qualitative comparisons of neural style transfer between the existing methods and the proposed MetaStyle using bilevel optimization. Arbitrary style transfer models observe neither the content images nor the style images during training.}
    \label{fig:compare}
\end{figure*}

We compare the proposed MetaStyle with existing methods \cite{gatys2016image,johnson2016perceptual,li2017universal,huang2017arbitrary,shen2018neural,sheng2018avatar,chen2016fast} in terms of speed, flexibility, and quality. Specifically, for these existing methods, we use the pre-trained models made publicly available by the authors. To adapt MetaStyle to a specific style, we train the MetaStyle model using only 200 iterations on MS-COCO dataset, which amounts to an additional 24 seconds of training time with a Titan Xp GPU. For Gatys \etal, we optimize the input using 800 update steps. For Chen \etal, we use its fast approximation model. All five levels of encoders and decoders are employed in our experiments involving Li \etal.

\textbf{Speed and Flexibility:}
Table~\ref{tbl:benchmark} summarizes the benchmarking results regarding style transfer speed and model flexibility. As shown in the table, our method achieves the same efficiency as Johnson \etal and Shen \etal. Additionally, unlike Shen \etal that introduces a gigantic parameter prediction model, MetaStyle is parsimonious with roughly the same number of parameters as Johnson \etal. While Johnson \etal requires training a new style model from scratch, MetaStyle could be immediately adapted to any style with a negligible number of updates under 30 seconds. This property significantly reduces the efforts in arbitrary style transfer and, at the same time, maintains a high image generation quality, as shown in the next paragraph.

\textbf{Quality:}
Figure~\ref{fig:compare} shows the qualitative comparisons of the style transfer between the existing methods and the proposed MetaStyle method. We notice that, overall, Gatys \etal and Johnson \etal obtain the best image quality among all the methods we tested. This observation coheres with our expectation, as Gatys \etal iteratively refines a single input image using an optimization method, whereas the model from Johnson \etal learns to approximate optimal solutions after seeing a large number of images and a fixed style, resulting in a better generalization.

Among methods capable of arbitrary style transfer, Li \etal applies style strokes excessively to the contents, making the style transfer results become deformed blobs of color, losing much of the image structures in the content images. Looking deep into Huang \etal, we notice that the arbitrary style transfer method produces images with unnatural cracks and discontinuities. Results from Shen \etal come with strange and peculiar color regions that likely result from non-converged image transformation models. Sheng \etal unnecessarily morphs the contours of the content images, making the generated artistic effects inferior. The inverse network from Chen \etal seems to apply the color distribution in the style image to the content image without successfully transferring the strokes and artistic effects in style.

\begin{figure*}[t!]
    \centering
    \captionsetup[subfloat]{farskip=0pt}
    \subfloat[Two-style interpolation results. The content image and style images are shown on the two ends.]{%
        \includegraphics[width=\linewidth]{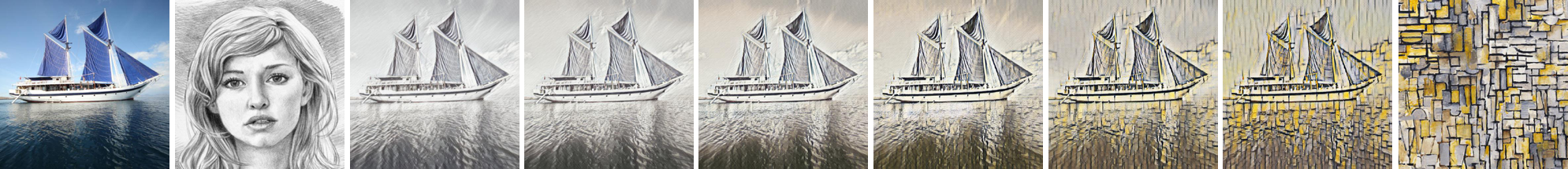}
        \label{fig:inter}
    }\\
    \subfloat[Video style transfer results. The left pane shows the style and the right pane contents and stylized video sequence.]{%
        \includegraphics[width=\linewidth]{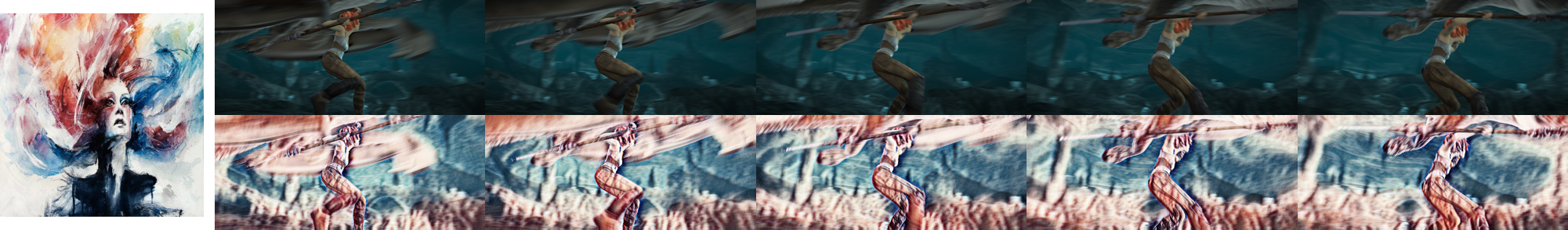}
        \label{fig:sequence}
    }
    \caption{Style interpolation and video style transfer.}
\end{figure*}

\begin{figure}[b!]
    \centering
    \captionsetup[subfloat]{farskip=0pt}
    \subfloat{%
        \begin{tabular}[b]{c}%
            \includegraphics[width=0.25\linewidth]{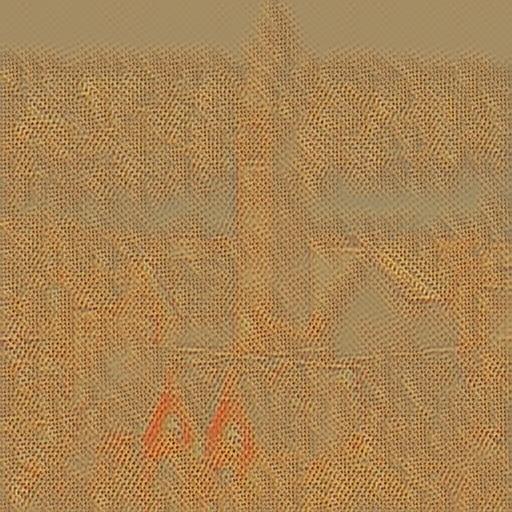}\\
            \includegraphics[width=0.25\linewidth]{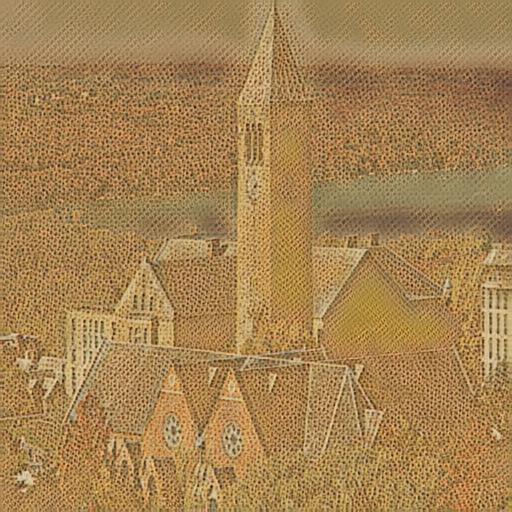}
        \end{tabular}}
    \subfloat{%
        \includegraphics[width=0.725\linewidth,height=4.5cm,trim={0.5cm 0.2cm 1cm 1cm},clip]{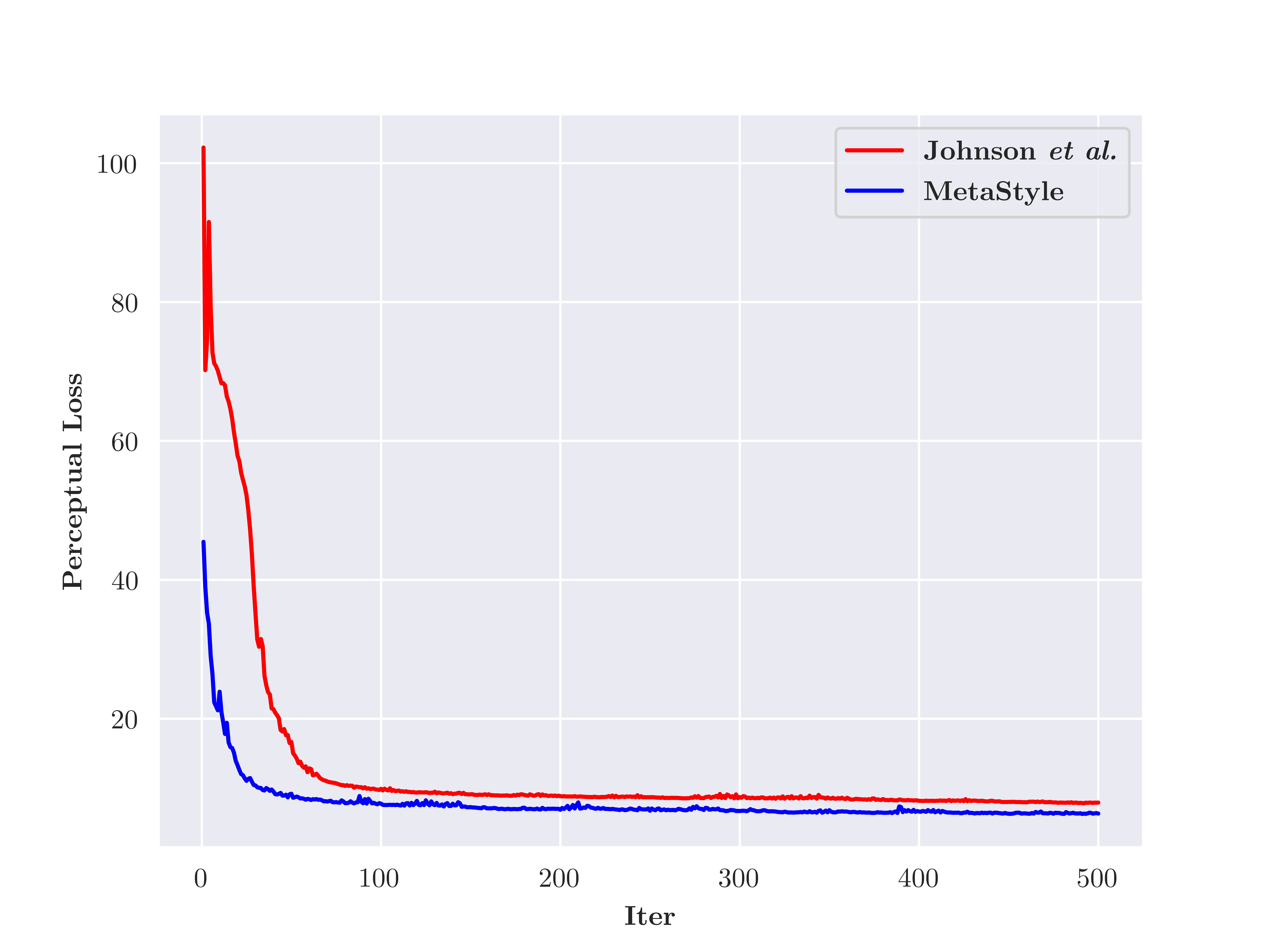}}
    \caption{Comparison with Johnson \etal. (Left) The results using (upper) Johnson \etal and (lower) the proposed MetaStyle. (Right) The perceptual loss during evaluation.}
    \label{fig:johnson}
\end{figure}

In contrast, MetaStyle achieves a right balance between styles and contents comparable to Johnson \etal. Such property should be attributed to the image transformation network shown in Johnson \etal \shortcite{johnson2016perceptual}, while the fast adaptation comes from our novel formulation; see next paragraph.

\textbf{Detailed Comparison with Johnson \etal \shortcite{johnson2016perceptual}:}
To show the fast adaptation rooted in our formulation, we train a fast approximation model from Johnson \etal and adapt our MetaStyle using the same number of updates on a shared style with the same learning rate. Figure~\ref{fig:johnson} shows the results after 200 training iterations and the curve for the perceptual loss during evaluation. It is evident that while Johnson \etal still struggles to figure out a well-balanced interpolation between the style manifold and the content manifold, MetaStyle could already generate a high-quality style transfer result with a good equilibrium between style and content. This contrast becomes even more significant considering that a fully trained model from Johnson \etal requires about 160,000 iterations and an adapted MetaStyle model only 200. The loss curve also shows consistently lower evaluation error compared to Johnson \etal, numerically proving the fast adaptation property of the proposed MetaStyle. %The relationship between the loss and the style transfer quality is still unresolved, but the loss difference around 1.5 does mean notable quality gap.

\begin{figure}[b!]
    \centering
    \captionsetup[subfloat]{farskip=0pt}
    \subfloat{%
        \begin{tabular}[b]{c}%
            \includegraphics[width=0.25\linewidth]{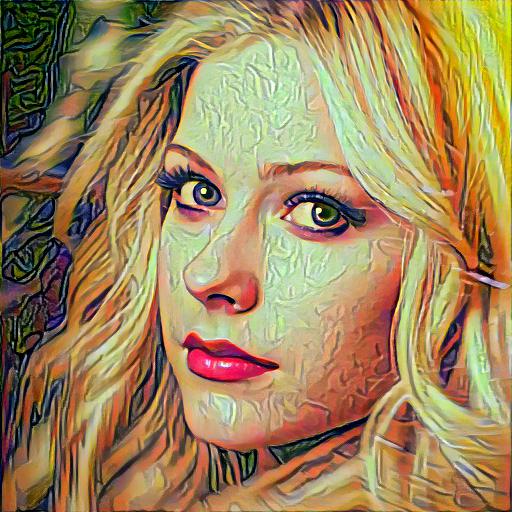}\\
            \includegraphics[width=0.25\linewidth]{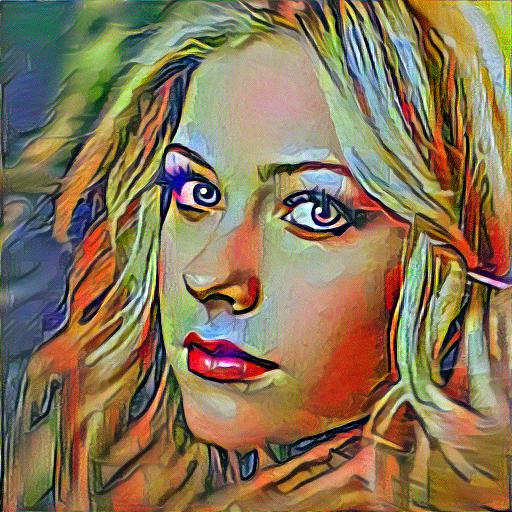}
        \end{tabular}}
    \subfloat{%
        \includegraphics[width=0.725\linewidth,height=4.5cm,trim={0.5cm 0.2cm 1cm 1cm},clip]{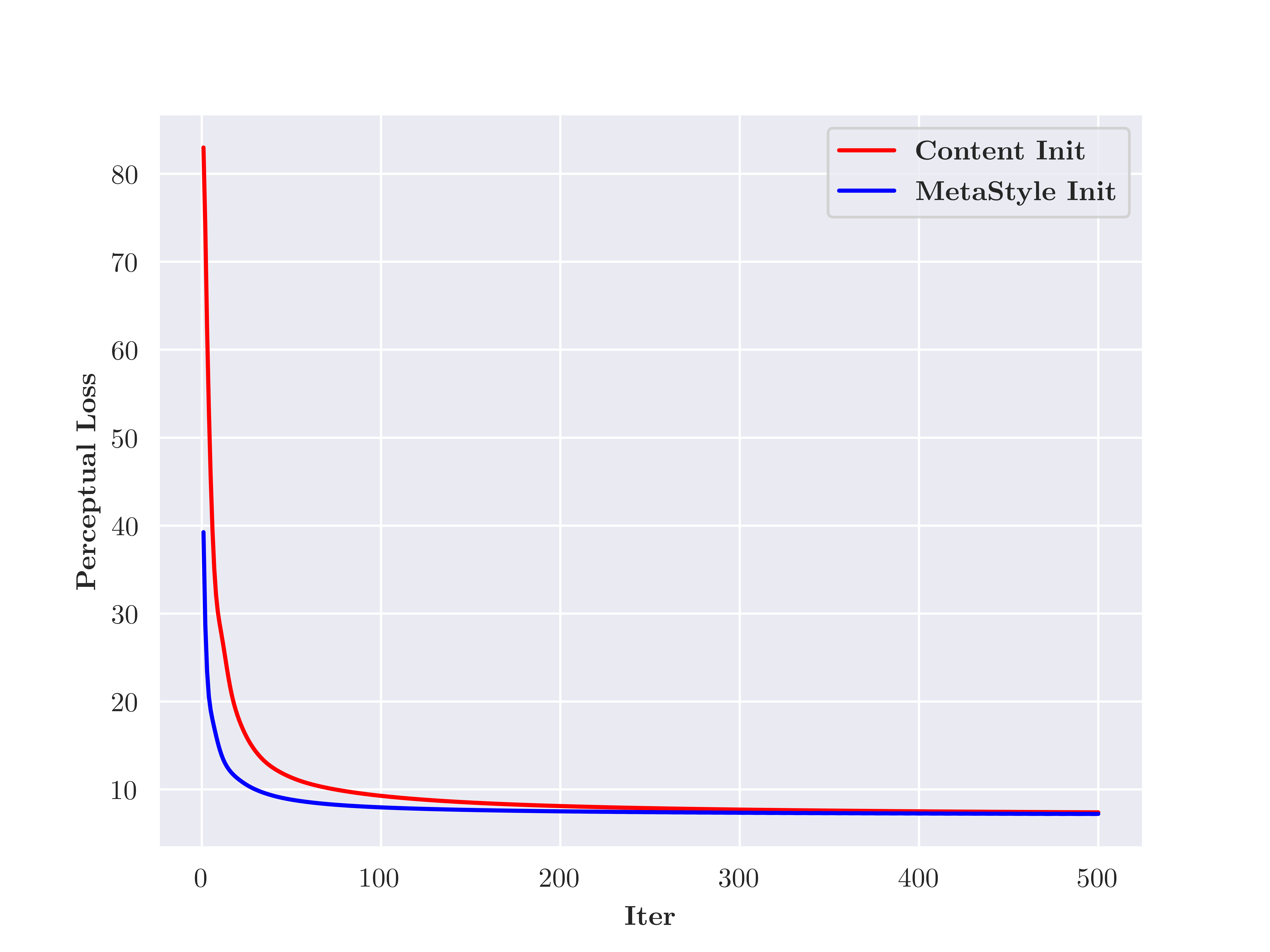}}
    \caption{Comparison with Gatys \etal. (Left) The results using (upper) Gatys \etal and (lower) the proposed MetaStyle. (Right) The perceptual loss.}
    \label{fig:gatys}
\end{figure}

\textbf{Detailed Comparison with Gatys \etal \shortcite{gatys2016image}:}
As mentioned in Section~\ref{sec:formulate}, MetaStyle, before adaptation, provides a style-neutral representation and serves as a better starting point for the optimization-based method. We empirically illustrate in Figure~\ref{fig:gatys}, in which we compare initializing the optimization with either the content image or the style-neutral representation. We notice that after 150 steps, Gatys \etal only starts to apply minor style strokes to the content while MetaStyle-initialized method could already produce a well-stylized result. Given that MetaStyle is not directly formulated to find a good starting point, this effect is surprising, showing the generalization ability of the representation discovered by the proposed MetaStyle.

\subsection{Additional Experiments}

We now present two additional experiments to demonstrate the style-neutral representation learned by MetaStyle.

\textbf{Style Interpolation:}
To interpolate among a set of styles, we perform a convex combination on the parameters of adapted MetaStyle models learned after 200 iterations. Figure~\ref{fig:inter} shows the results of a two-style interpolation. 

\textbf{Video style transfer:}
We perform the video style transfer by first training the MetaStyle model for 200 iterations to adapt to a specific style, and then applying the transformation to a video sequence frame by frame. Figure~\ref{fig:sequence} shows the video style transfer results in five consecutive frames. Note that our method does not introduce the flickering effect that harms aesthetics. Additional videos are provided in the supplementary files.

\section{Conclusion}

In this paper, we present the MetaStyle, which is designed to achieve a right three-way trade-off among speed, flexibility, and quality in neural style transfer. Unlike previous methods, MetaStyle considers the arbitrary style transfer problem in a new scenario where a small (even negligible) number of post-processing updates are allowed to adapt the model quickly to a specific style. We formulate the problem in a novel bilevel optimization framework and solve it using \ac{maml}. In experiments, we show that MetaStyle could adapt quickly to an arbitrary style within 200 iterations. Each adapted model is an image transformation network and benefits the high efficiency and style transformation quality on par with Johnson \etal. The detailed comparison and additional experiments also show the generalized style-neutral representation learned by MetaStyle. These results show MetaStyle indeed achieves a right trade-off.%among speed, flexibility, and quality.

\noindent\textbf{Acknowledgments:}
The work reported herein was supported by the International Center for AI and Robot Autonomy (CARA).

% {\fontsize{8.7pt}{8.7pt} \selectfont
% \bibliography{main.bbl}
% \bibliographystyle{aaai}
% }

\end{document}